\documentclass{article}

\usepackage{amsmath,amssymb,amsfonts}
\usepackage{booktabs}
\usepackage{graphicx}
\usepackage{xcolor}
\usepackage{enumitem}
\usepackage{float}
\usepackage{placeins}
\usepackage{algorithm}
\usepackage{algorithmic}
\usepackage{wrapfig}
\usepackage{uxbench}
\usepackage[hypcap=false]{caption}
\usepackage{subcaption}
\usepackage{xurl}

\newcommand{\preprintshorttitle}{Teach to Learn: Hint Annealing for Self-improving LLM Reasoning}
\makeatletter
\patchcmd{\headrule}
  {UXBench: Benchmarking User Experience in AI Assistants}
  {\preprintshorttitle}
  {}{\PackageError{hatch}{Could not replace the template running title}{}}
\patchcmd{\@maketitle}
  {\begin{center}
\vspace{-1cm}
\begin{tabular}{rl}
\yuanbao & \url{\homelink}\\
\end{tabular}
\vspace{0.2cm}
\end{center}}
  {}{\relax}{\PackageError{hatch}{Could not remove the template project link}{}}
\makeatother

\title{Teach to Learn: Hint Annealing for Self-improving LLM Reasoning}
\author{
\bfseries The Hong Kong University of Science and Technology \quad \mbox{Yuanbao Team, Tencent}\thanks{\protect\hypertarget{Hfootnote.1}{}Full author list at the end of the paper.}
}
\hypersetup{
  pdftitle={Teach to Learn: Hint Annealing for Self-improving LLM Reasoning},
  pdfauthor={Zile Wang, Zijian Li, Haodong Wang, Jian Liu, Qianli Liu, Lucas Muli, Blaze Chen, Song Guo}
}

\begin{document}
\maketitle

\begin{abstract}
Group Relative Policy Optimization (GRPO) improves language-model reasoning by comparing verified rewards among multiple solution rollouts for each query. However, difficult training queries can yield only incorrect rollouts, leaving GRPO with no reward contrast or learning signal.
Prior hint-based methods construct auxiliary hints from solution evidence and use them to re-solve failed queries, recovering learning signal. Yet the resulting trajectories are typically treated as ordinary solution trajectories despite being generated under an assisted condition unavailable at evaluation.
We discover \emph{hinted reward shift}: recovered reward contrast can concentrate policy updates on hinted trajectories, limiting improvement without hints. This also creates a trade-off: increasing hinted trajectories can accelerate early learning but intensify reward shift later.
To address this problem, we propose \textsc{HATCH} (\emph{Hint-Annealed Self-Teaching}), an online single-policy framework that learns from both generating and using its own hints to improve reasoning without assistance.
To mitigate hinted reward shift, we introduce online weighting to anneal the contribution of hinted trajectories. However, learning to generate hints can conflict with improving query solving. We therefore use gradient projection to remove the opposing component of hint-generation updates. Together, these designs support self-improvement by enabling the policy to create learning opportunities for itself and turn them into stronger reasoning without hints.
We evaluate our method on mathematical reasoning benchmarks and outperform state-of-the-art methods by 1.02~pp on Llama-3.2-1B-Instruct, 2.84~pp on Qwen3-1.7B, and 4.32~pp on Qwen3-8B.
\end{abstract}

\section{Introduction}

Reinforcement learning with verifiable rewards (RLVR) improves reasoning in
language models by training a policy model with automatically checked
answers~\citep{shao2024deepseekmath,yu2026dapo}. For each query, the policy
model samples several solution rollouts. An automatic verifier determines
whether each final answer is correct and assigns a reward accordingly. Group
Relative Policy Optimization (GRPO) then compares these rewards within the
rollout group and converts their differences into relative advantages: responses that perform
better than their group peers are reinforced, while worse responses are
suppressed. Thus, under binary correctness rewards, those groups containing both successful
and unsuccessful rollouts provide the reward contrast needed to specify a local
direction for improving the policy.
However, queries within a batch vary in difficulty. For queries that the
current policy rarely solves, a limited set of sampled rollouts can all be
incorrect, leaving no reward contrast. As a result, these identical rewards yield zero
relative advantages, leaving the policy without learning signal from queries it
still struggles to solve.

Existing work addresses this issue along two complementary dimensions:
improving rollout budget allocation and recovering learning signal from failed queries.
The first line of work improves the use of the rollout budget through adaptive
rollout allocation, selective rollout sampling, and dynamic
filtering~\citep{li2025knapsack,qu2025optimizing,zheng2026act,yu2026dapo}.
These strategies improve how computation is allocated and rollouts are
selected for updates based on reward contrast already exposed by
original rollouts.
The second line of work constructs hints to recover learning signal from failed
queries. 
These methods alter the condition under which difficult queries are solved by
supplying partial reasoning, solution-derived hints, or learned guidance~\citep{li2025questa,zhang2026scafgrpo,chen2025nudging,liao2026self,xia2026learning}.
By making successful solution paths easier to discover, this assistance can
turn all-incorrect original rollout groups into re-solves with both successes
and failures, thereby recovering reward contrast.
Yet prior methods focus primarily on constructing hints, while the resulting
trajectories are optimized under an assisted condition and can shift learning
away from original-query solving.

Our analysis reveals that reward contrast recovered under hints can make the shared update
favor hinted trajectories over the original-query trajectories used at evaluation. 
When original rollouts provide no reward contrast, hinted re-solving
can recover learning signal, but the resulting updates directly
optimize solving under hints.
As examined in Section~\ref{sec:preliminaries}, hinted and original
solving can induce substantially different gradient directions,
even when both conditions yield reward contrast.
Sustained hinted feedback can therefore continue to optimize
solving under hints without producing corresponding gains on
the original query.
We call this phenomenon \emph{hinted reward shift}. Hint-based RL must therefore
not only recover learning signal from failed queries, but also prevent the shared update from favoring solving with hints
over original-query solving.

\begin{figure*}[t]
\centering
\includegraphics[width=\textwidth]{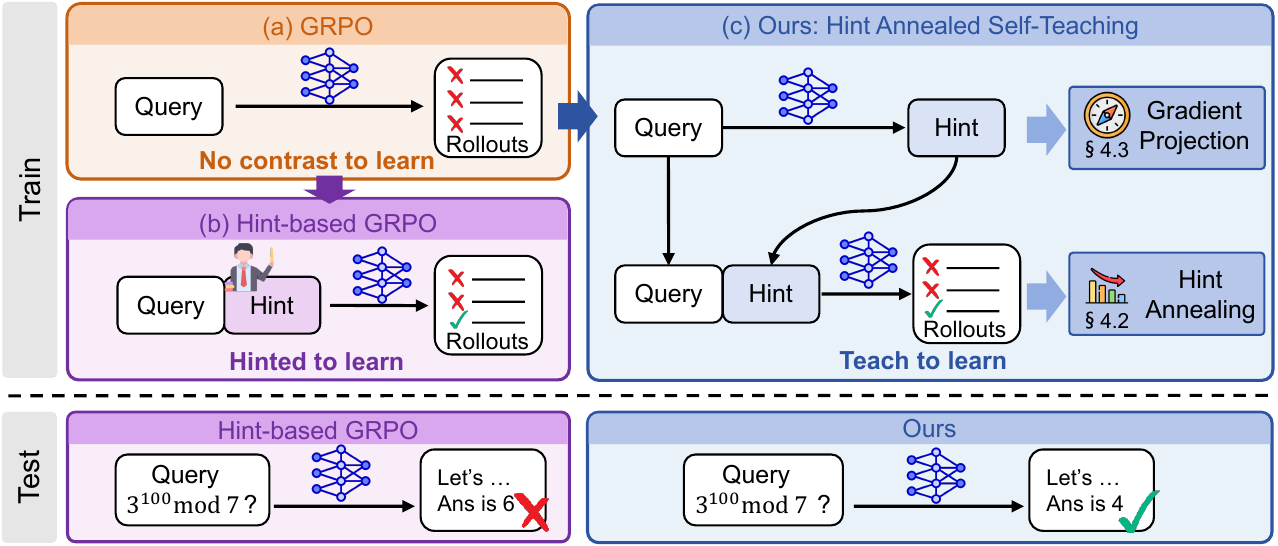}
\caption{Comparison between GRPO, hint-based GRPO, and ours.
(a) GRPO receives no group-relative learning signal for hard queries with all-incorrect rollouts. (b) Hint-based GRPO restores learning signal by re-solving under a hint, yielding strong improvement during training but limited improvement during testing. (c) Ours learns from generating and using its own hints, while gradually shifting training toward the original solving ability without hints.}
\label{fig:intro-overview}
\end{figure*}

To address this challenge, we propose \textsc{HATCH}
(\emph{Hint-Annealed Self-Teaching}), which frames self-improvement as a
teach-to-learn process: the policy uses self-generated hints to turn its own
failures into learnable re-solves, and learns from both the resulting solutions
and the effectiveness of its teaching (Figure~\ref{fig:intro-overview}). 
Meanwhile, hints should serve as an annealing
assistance: hinted-solve feedback is most valuable while unassisted rollouts
provide little reward contrast and should recede as original solving improves.
We realize this annealing through data-driven online weighting.
Yet hint-generation serves a distinct reasoning role from solving, so the gradient signals need not point in the same direction.
We address this conflict through gradient projection, removing the
hint-generation component that opposes solution updates. 
Together, these designs enable the policy to guide its own learning
toward sustained self-improvement in unassisted reasoning.
Our contributions are:

\begin{itemize}[leftmargin=1.45em]
  \item We discover that learning signal recovered under hints can favor hinted solving. We term this \emph{hinted reward shift}, revealing a trade-off between early gains from hint recovery and sustained improvement on original query solving.

  \item We propose an online self-teaching framework that jointly learns
  hint generation and query solving with two components: online weighting
  to regulate the contribution of hinted trajectories and gradient projection
  to coordinate hint-generation learning with query solving.

  \item We evaluate our method across nine mathematical reasoning benchmarks and deliver
  consistent accuracy gains over state-of-the-art methods across three model
  scales: 1.02~pp on Llama-3.2-1B-Instruct, 2.84~pp on Qwen3-1.7B,
  and 4.32~pp on Qwen3-8B.
\end{itemize}

\section{Related Work}

\noindent\textbf{RL Signal Utilization.}
GRPO derives learning signal from within-group reward contrast, while difficult queries yielding only incorrect rollouts provide no
relative preference signal.
To make better use of existing learning signal
under the original rollout condition, prior work acts in three complementary
ways. Sampling methods assign heterogeneous rollout budgets across prompts or predict
prompt difficulty to prioritize those likely to yield informative gradients
\citep{li2025knapsack,qu2025optimizing}.
Filtering and signal-shaping methods retain groups at appropriate difficulty,
skip prompts predicted to be uninformative, down-sample redundant rollouts, or
construct advantages for zero-variance groups
\citep{bae2026online,zheng2026act,yu2026dapo,xu2026not,le2026no}.
Curriculum methods schedule tasks from easy to hard or adapt the task mixture
according to the policy's evolving learning progress
\citep{bengio2009curriculum,chen2025self,parashar2026curriculum}.
Together, these approaches improve the allocation of computation and updates
under the original rollout condition. They are complementary to our work, which
uses solution-derived hints to recover learning signal from queries that remain
uninformative under original rollouts.

\noindent\textbf{Hint-Based Signal Recovery.}
On the other hand, hint-based RL recovers learning signal from difficult queries by re-solving
them under auxiliary contexts. 
One line of work explores auxiliary contexts that make difficult queries more
accessible through partial solutions, tiered scaffolds, stepwise reasoning
prefixes, or selected knowledge guidance
\citep{li2025questa,zhang2026scafgrpo,zhang2026stephint,yu2026knowrl}.
Another form of such auxiliary context guides exploration closer to successful
reasoning paths through oracle prefixes or expert-anchored rollouts
\citep{qu2026pope,zhang2026bread}.
A second line studies how useful hints can be generated and maintained during
training through self-generated abstractions, privileged solution-derived
hints, or learned hinters conditioned on current policy failures
\citep{chen2025nudging,liao2026self,xia2026learning}. Within this direction,
external guidance can further be kept compatible with online policy updates
through abstract meta-hints and affinity-aware optimization
\citep{wang2026don}.
Together, these methods establish how auxiliary contexts
can recover reward contrast and how useful hints can be produced. 
However, effective hints can still induce auxiliary updates that favor
hinted solving over solving without hints. We instead study how learning to provide guidance can itself improve
a policy's unassisted reasoning.

\section{Observation}
\label{sec:preliminaries}

\subsection{Preliminaries}
\noindent\textbf{GRPO and Hint-Based Optimization.}
Given verifiable outcome rewards, GRPO samples \(G\) responses
\(\{y_{i,k}\}_{k=1}^{G}\) for each query
\(q_i\in\mathcal{B}_t\) at each update step \(t\), where \(i\) indexes
individual queries and \(k\) indexes sampled responses within each query group.
The verifier \(V\) assigns each response the outcome reward
\(R_{i,k}=V(q_i,y_{i,k})\). GRPO then computes the group-relative advantage
\begin{equation}
A_{i,k}
=
\frac{R_{i,k}-\frac{1}{G}\sum_{k'=1}^{G}R_{i,k'}}
{\operatorname{Std}\!\left(\{R_{i,k'}\}_{k'=1}^{G}\right)+\epsilon}.
\label{eq:grpo-advantage}
\end{equation}
This advantage is assigned to the tokens of \(y_{i,k}\). GRPO minimizes the
standard clipped objective
\begin{equation}
\mathcal{L}_{\mathrm{GRPO}}(\theta)
=
-\mathbb{E}_{q_i\sim\mathcal{B}_t}\left[
\frac{1}{G}\sum_{k=1}^{G}
\ell_{\mathrm{clip}}\!\left(y_{i,k}, A_{i,k}\right)
\right],
\label{eq:grpo-objective}
\end{equation}
where \(\ell_{\mathrm{clip}}\) denotes the standard clipped surrogate~\citep{schulman2017proximal} for each
sampled response.
We define query groups with zero, some, or all correct rollouts as
\emph{solve-none}, \emph{solve-partial}, or \emph{solve-all}, denoted by
\(\mathcal{B}_t^{\mathrm{none}}\),
\(\mathcal{B}_t^{\mathrm{partial}}\), and
\(\mathcal{B}_t^{\mathrm{all}}\), respectively. Hint-based methods re-solve queries in \(\mathcal{B}_t^{\mathrm{none}}\)
under generated hints to recover reward contrast.
Let \(\mathcal{Y}\) and \(\mathcal{Y}^{h}\) denote original and hinted
trajectories with advantages \(A^{\mathrm{orig}}\) and
\(A^{\mathrm{hinted}}\). Applying the GRPO objective in
Eq.~\eqref{eq:grpo-objective} we have
\begin{equation}
\mathcal{L}_{\mathrm{hint\mbox{-}based}}(\theta)
=
\underbrace{
\mathcal{L}_{\mathrm{GRPO}}\!\left(
\mathcal{Y}, A^{\mathrm{orig}};\theta\right)
}_{\mathcal{L}_{\mathrm{solve}}}
+
\underbrace{
\mathcal{L}_{\mathrm{GRPO}}\!\left(
\mathcal{Y}^{h}, A^{\mathrm{hinted}};\theta\right)
}_{\mathcal{L}_{\mathrm{hinted\mbox{-}solve}}}.
\label{eq:hint-based-losses}
\end{equation}

Existing work has primarily focused on how to construct effective hints: by
generating compact abstractions from solution, revealing privileged
intermediate reasoning, or training a separate hinter to produce guidance for
a subsequent attempt~\citep{chen2025nudging,liao2026self,xia2026learning}.

\begin{figure*}[t]
\centering
\begin{minipage}[b]{0.325\textwidth}
  \centering
  \includegraphics[width=\linewidth]{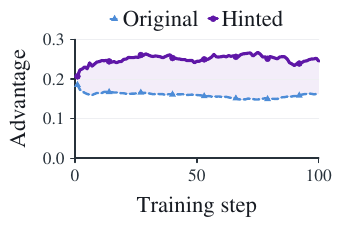}\\[-0.25em]
\end{minipage}\hfill
\begin{minipage}[b]{0.325\textwidth}
  \centering
  \includegraphics[width=\linewidth]{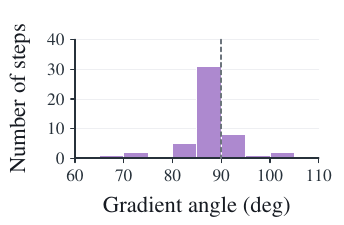}\\[-0.25em]
\end{minipage}\hfill
\begin{minipage}[b]{0.325\textwidth}
  \centering
  \includegraphics[width=\linewidth]{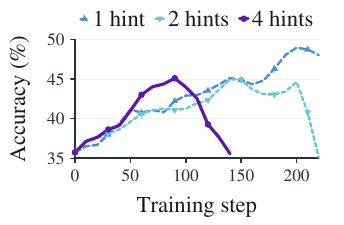}\\[-0.25em]
\end{minipage}
\caption{
Evidence of hinted reward shift on Qwen3-8B.
(a) Left: Mean absolute advantages of original and hinted trajectories.
(b) Middle: Gradient-angle distribution between original and hinted solving
over steps 1--50.
(c) Right: Original-query accuracy with different numbers of hints.
}
\label{fig:hinted-reward-shift}
\end{figure*}

\subsection{Hinted Trade-off Problem}
\noindent\textbf{Hint-Assisted Contrast Recovery Leads to Reward Shift.}
Figure~\ref{fig:hinted-reward-shift}(a--b) examines the learning
signals of the two objectives in Eq.~\eqref{eq:hint-based-losses}
through their mean absolute advantages and gradient directions.
Over training, hinted-solve trajectories retain mean absolute advantages
comparable to or larger than those of original-solve trajectories,
while the two objectives often produce nearly orthogonal gradients.
Such advantage comparison reflects an asymmetry: queries in
\(\mathcal{B}_t^{\mathrm{none}}\) provide no group-relative learning signal
through original rollouts, while re-solving them under hints can
constantly recover reward contrast.
Because this recovered signal directly optimizes solving under hints, its
sustained contribution can favor
\(\mathcal{L}_{\mathrm{hinted\mbox{-}solve}}\) over
\(\mathcal{L}_{\mathrm{solve}}\) in policy optimization.
We call this tendency for learning signal to concentrate on hinted solving
\emph{hinted reward shift}.
These observations suggest that sustained reward contrast under hints
can lead to favoring hinted solving without ensuring corresponding gains
on the original query.

\noindent\textbf{Trade-off Between Early Gains and Hinted Reward Shift.}
Given this shift, we examine how policy improvement and the shift evolve
under stronger hinted learning signals. To this end, we increase the
number of hints per failed query, adding more hinted-solve groups to
\(\mathcal{L}_{\mathrm{hinted\mbox{-}solve}}\).
Figure~\ref{fig:hinted-reward-shift}(c) shows that multiple hints yield faster
early accuracy gains but subsequently deteriorate, whereas the single-hint
setting continues to improve at a slower pace. This reflects the changing availability of learning signal
from original rollouts: when
\(\mathcal{B}_t^{\mathrm{none}}\) is large, hinted-solve groups
turn zero-signal queries into informative comparisons; but their continued influence may limit later
improvement as original learning signal emerges. Therefore, hinted trajectories should contribute most when many queries lack
learning signal and recede as that signal emerges, allowing the model to retain
its early gains.
This observation suggests a trade-off between faster early gains
and sustained original-query improvement, consistent with
\emph{hinted reward shift}. An effective method should therefore
fully exploit the learning potential
of hinted trajectories while keeping
the ability to solve without hints.

\section{Method}\label{sec:method}
\begin{figure*}[t]
\centering
\includegraphics[width=\textwidth]{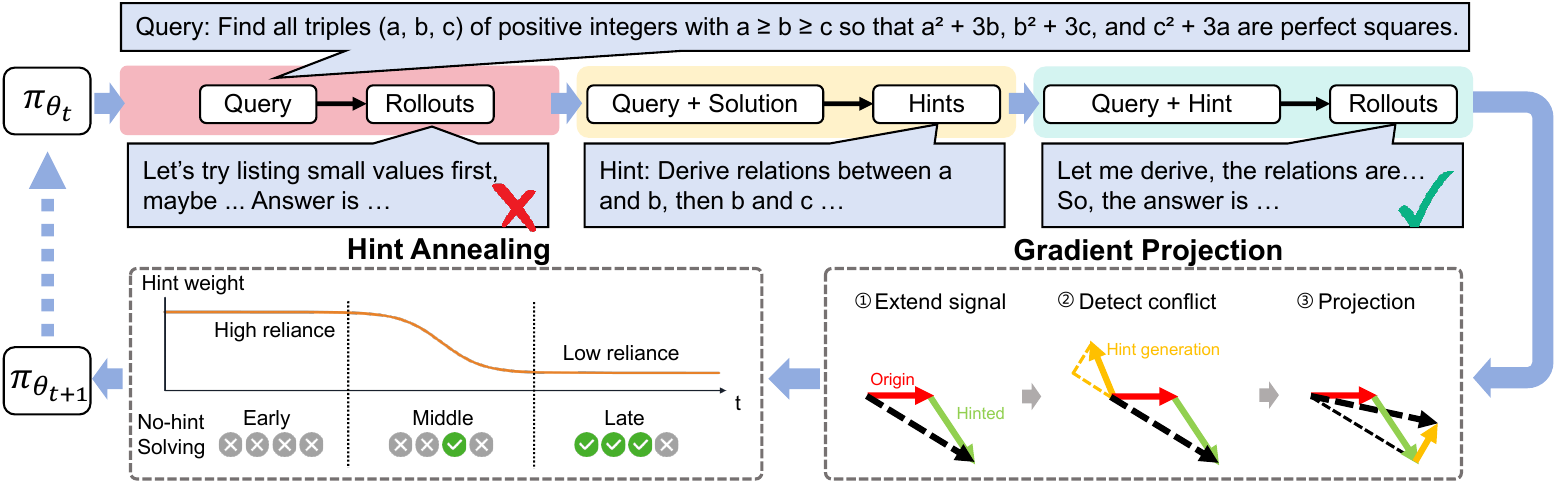}
\caption{Overview of our method. For solve-none queries, it first self-generates hints from solutions and then re-solves the query with the hints, yielding three complementary training trajectories: native solving, hint generation, and hinted solving. Online weighting adaptively controls the contribution of hinted-solving feedback, while gradient projection removes conflicting components from hint-generation updates. The resulting signals are jointly used to update a single shared policy.}
\label{fig:method-overview}
\end{figure*}

To address the challenge in Section~\ref{sec:preliminaries},
we propose \textsc{HATCH} (\emph{Hint-Annealed Self-Teaching}), an online
single-policy framework (Figure~\ref{fig:method-overview}). It constructs hint-generation and
hinted-solving objectives from initially failed queries in
Section~\ref{sec:self-hints}, regulates the contribution of hinted
trajectories through online weighting in
Section~\ref{sec:adaptive-integration}, and uses gradient projection to
address role divergence in Section~\ref{sec:alignment}.

\subsection{Online self-hinting for valuable trajectories}
\label{sec:self-hints}

Section~\ref{sec:preliminaries} shows the potential of additional hinted
trajectories to improve policy learning. To continually generate more useful
hints, we jointly train hint generation and query solving within one policy,
using re-solving outcomes to supervise hint generation.
This introduces \(\mathcal{L}_{\mathrm{hint\mbox{-}gen}}\) alongside
\(\mathcal{L}_{\mathrm{solve}}\) and
\(\mathcal{L}_{\mathrm{hinted\mbox{-}solve}}\) in
Eq.~\eqref{eq:hint-based-losses}.
Specifically, for each query \(q_i\in\mathcal{B}_t^{\mathrm{none}}\),
we condition the policy on its reference solution \(z_i\) to sample
\(M\) hints as
\begin{equation}
h_{i,j}
\sim \pi_t\!\left(
\cdot\mid P_{\mathrm{hint}}(q_i,z_i)\right),
\qquad
j=1,\ldots,M.
\end{equation}
To evaluate each hint through its effect on solving, we append \(h_{i,j}\)
to \(q_i\) and sample \(G\) re-solves
\(\{y^h_{i,j,k}\}_{k=1}^{G}\) from the same policy.
We score these responses and compute
\(A^{\mathrm{hinted}}_{i,j,k}\) within each hinted-solve group using
Eq.~\eqref{eq:grpo-advantage}.
The same outcomes also supervise hint generation through the reward
\begin{equation}
R^{\mathrm{hint\mbox{-}gen}}_{i,j}
=
\mathcal{R}_{\mathrm{hint}}\!\left(
\Delta\mathrm{Pass}_{i,j},
\Delta\log p_{i,j}
\right).
\end{equation}
Here, \(\Delta\mathrm{Pass}_{i,j}\) measures the rollout accuracy gain,
and \(\Delta\log p_{i,j}\) measures the log-probability difference of the
successful trajectory under the hinted and original
conditions~\citep{xia2026learning}.
Applying Eq.~\eqref{eq:grpo-advantage} across the \(M\) hint rewards
yields \(A^{\mathrm{hint\mbox{-}gen}}_{i,j}\).
Using these advantages to optimize the hint-generation trajectories with
Eq.~\eqref{eq:grpo-objective} defines
\(\mathcal{L}_{\mathrm{hint\mbox{-}gen}}\).
We retain all \(M\) hinted-solve groups rather than selecting one.
Let \(T_O\), \(T_S\), and \(T_H\) denote the numbers of active response
tokens in original solving, hinted solving, and hint generation,
respectively, with \(T_F=T_O+T_S+T_H\). Each loss is averaged over the active response tokens
of its own trajectory.
The joint objective is
\begin{equation}
\mathcal{L}_{\mathrm{joint}}
=
\frac{T_O}{T_F}\mathcal{L}_{\mathrm{solve}}
+
\frac{T_S}{T_F}\mathcal{L}_{\mathrm{hinted\mbox{-}solve}}
+
\frac{T_H}{T_F}\mathcal{L}_{\mathrm{hint\mbox{-}gen}}.
\label{eq:joint-objective}
\end{equation}
As the policy learns to solve queries, its re-solving outcomes also refine
the hints it generates for its remaining failures.
The following subsections regulate the contribution of hinted
trajectories and the direction of hint-generation updates so that both
support learning to solve the original queries.

\subsection{Online Weighting of Hinted Trajectories}
\label{sec:adaptive-integration}
To mitigate hinted reward shift in the joint training of
Section~\ref{sec:self-hints}, we adjust the contribution of
\(\mathcal{L}_{\mathrm{hinted\mbox{-}solve}}\) according to the learning
signal available from original rollouts.
The relative proportions of unsolved groups and groups with reward contrast
indicate how much training can rely on original rollouts.
Let \(\rho_{\mathrm{none}}^{(t)}\) and
\(\rho_{\mathrm{partial}}^{(t)}\) denote
\(\lvert\mathcal{B}_t^{\mathrm{none}}\rvert/\lvert\mathcal{B}_t\rvert\)
and
\(\lvert\mathcal{B}_t^{\mathrm{partial}}\rvert/\lvert\mathcal{B}_t\rvert\),
respectively.
We smooth these proportions with Exponential Moving Averages (EMA) to determine the weight \(\beta_t\) as
\begin{equation}
p_t
=
\frac{\operatorname{EMA}(\rho_{\mathrm{none}}^{(t)})}
{\operatorname{EMA}(\rho_{\mathrm{none}}^{(t)})
+\operatorname{EMA}(\rho_{\mathrm{partial}}^{(t)})+\epsilon},
\qquad
\beta_t
=
\operatorname{clip}\!\left(\kappa p_t^\gamma,0,1\right).
\label{eq:online-weighting}
\end{equation}
As the balance shifts from \(\mathcal{B}_t^{\mathrm{none}}\) toward
\(\mathcal{B}_t^{\mathrm{partial}}\), \(\beta_t\) decreases, reducing
reliance on hinted trajectories as original reward contrast emerges.
The scale \(\kappa\) controls the overall strength, while \(\gamma\)
controls how sharply the weight decreases.
We apply \(\beta_t\) only to hinted-solve advantages as
\begin{equation}
\widetilde A^{\mathrm{hinted}}_{i,j,k}
=
\beta_t A^{\mathrm{hinted}}_{i,j,k}.
\label{eq:adaptive-hinted-credit}
\end{equation}
Online weighting thus mitigates hinted reward shift by letting hinted
trajectories supplement learning when original reward contrast is scarce,
while reducing their contribution as that contrast emerges.
Still, the direction of hint-generation updates requires separate treatment.

\subsection{Gradient Projection for Hint Generation}
\label{sec:alignment}
\edef\figfourparskip{\the\parskip}
\noindent
\begin{minipage}[t]{0.60\textwidth}
\setlength{\parskip}{\figfourparskip}
Although hint generation is rewarded by re-solving outcomes, its updates
need not support query solving within the shared policy.
Figure~\ref{fig:hint-gradient-angle} shows that the angle between
hint-generation gradient and the combined gradient from original-solve
and weighted hinted-solve trajectories fluctuates around \(90^\circ\)
and frequently exceeds it, indicating a learning divergence.
We therefore use gradient projection to make hint-generation updates
compatible with the update induced by solution trajectories.

Let \(\mathcal{L}_{\mathrm{full}}\) denote the joint loss of all three
objectives after applying Eq.~\eqref{eq:adaptive-hinted-credit}.
We then compute its full gradient and the hint-generation gradient
separately as
\end{minipage}\hfill
\begin{minipage}[t]{0.36\textwidth}
\vspace{0pt}
\centering
\includegraphics[
  width=\linewidth,
  trim=0 0 0 10bp,
  clip
]{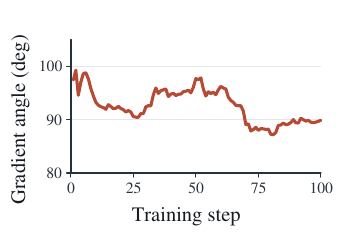}
\captionof{figure}{\raggedright
Angle between the hint-generation gradient and the solution-side gradient.}
\label{fig:hint-gradient-angle}
\end{minipage}
\par
\begin{equation}
g_F
=
\nabla_\theta\mathcal{L}_{\mathrm{full}},
\qquad
g_H
=
\nabla_\theta\mathcal{L}_{\mathrm{hint\mbox{-}gen}}.
\label{eq:gradient-definition}
\end{equation}
Since \(g_H\) is normalized over hint-generation tokens alone,
its contribution to the full gradient across all three objectives
is scaled by \(T_H/T_F\) as
\begin{equation}
g_H^\prime
=
\frac{T_H}{T_F}g_H,
\qquad
g_P
=
g_F-g_H^\prime.
\end{equation}
Here, \(g_P\) combines the gradients from the solution side trajectories as the reference direction.
When \(\langle g_H^\prime,g_P\rangle\) is nonnegative, hint-generation
learning does not oppose the solution-side update and is retained unchanged.
Otherwise, we remove only its component opposite to \(g_P\) as
\begin{equation}
g_{\mathrm{update}}
=
g_P+
\begin{cases}
g_H^\prime-
\dfrac{\langle g_H^\prime,g_P\rangle}
{\lVert g_P\rVert_2^2+\epsilon}g_P,
&
\langle g_H^\prime,g_P\rangle<0,\\[10pt]
g_H^\prime,
&
\text{otherwise}.
\end{cases}
\label{eq:hint-projection}
\end{equation}
The projection preserves \(g_P\) and the non-opposing component of
hint-generation learning. This allows the policy to learn useful hints
without counteracting the update from solution trajectories.

\begin{algorithm}[b]
\caption{\emph{Hint-Annealed Self-Teaching}}
\label{alg:online-self-hinting}
\begin{algorithmic}[1]
\STATE \textbf{Input:} rollout policy \(\pi_t\), training batch
\(\{(q_i,z_i)\}\)
\STATE Sample \(G\) original-solve responses for each \(q_i\); compute
\(R_{i,k}\), \(A^{\mathrm{orig}}_{i,k}\), and
\(\mathcal{B}_t^{\mathrm{none}}\)
\FOR{each \(q_i\in\mathcal{B}_t^{\mathrm{none}}\)}
    \STATE Sample \(M\) hints \(h_{i,j}\) conditioned on \(q_i,z_i\)
    \STATE For each hint, sample \(G\) hinted-solve responses
    \STATE Compute \(A^{\mathrm{hinted}}_{i,j,k}\) and
    \(A^{\mathrm{hint\mbox{-}gen}}_{i,j}\) from the re-solve outcomes
\ENDFOR
\STATE Compute \(\beta_t\) using Eq.~\eqref{eq:online-weighting} and obtain
\(\widetilde A^{\mathrm{hinted}}_{i,j,k}\) using
Eq.~\eqref{eq:adaptive-hinted-credit}
\STATE Form \(\mathcal{L}_{\mathrm{full}}\), compute
\(g_F,g_H,g_H^\prime,g_P\), and obtain \(g_{\mathrm{update}}\) using
Eq.~\eqref{eq:hint-projection}
\STATE Update \(\theta_{t+1} \leftarrow
\operatorname{Optimizer}(\theta_t,g_{\mathrm{update}})\)
\end{algorithmic}
\end{algorithm}

\subsection{Hint Annealed Self Teaching}
\label{sec:unified-update}
Having specified online weighting and gradient projection, we now combine
the three objectives in a single policy update.
Let \(\widetilde{\mathcal{L}}_{\mathrm{hinted\mbox{-}solve}}\) denote
the hinted-solve loss evaluated with the weighted advantages
\(\widetilde A^{\mathrm{hinted}}\) from
Eq.~\eqref{eq:adaptive-hinted-credit}.
The joint loss before gradient projection is
\begin{equation}
\mathcal{L}_{\mathrm{full}}(\theta)
=
\frac{T_O}{T_F}\mathcal{L}_{\mathrm{solve}}
+
\frac{T_S}{T_F}\widetilde{\mathcal{L}}_{\mathrm{hinted\mbox{-}solve}}
+
\frac{T_H}{T_F}\mathcal{L}_{\mathrm{hint\mbox{-}gen}}.
\label{eq:full-objective}
\end{equation}
We apply Eq.~\eqref{eq:hint-projection} to the hint-generation component
of \(\nabla_\theta\mathcal{L}_{\mathrm{full}}\) and use the resulting
\(g_{\mathrm{update}}\) to update the policy.
Algorithm~\ref{alg:online-self-hinting} summarizes one training update.
At inference, the policy receives only the original query, without hints
or reference solutions.

\begin{table}[t]
\caption{
Per-dataset no-hint accuracy (\%) for the initial models and each
trained method. Bold marks the best result and underlining marks the second-best
distinct result with ties retained.
}
\label{tab:main-results}
\begin{center}
\scriptsize
\setlength{\tabcolsep}{2.5pt}
\resizebox{\linewidth}{!}{\begin{tabular}{lcccccccc}
\toprule
Method & Math500 & Minerva & Oly. & AIME24/25/26 & AMC23 & HMMT25 & BRUMO25 & Avg. \\
\midrule
\textbf{Llama-3.2-1B-Instruct} & 24.04 & 4.41 & 4.71 & 1.11/0.00/0.00 & 5.00 & 0.00 & 1.11 & 4.49 \\
+ GRPO & 25.45 & 5.02 & 4.71 & \textbf{2.22}/\underline{1.11}/\underline{1.11} & 6.67 & \underline{1.11} & 1.11 & 5.39 \\
+ QuESTA & \underline{26.92} & \underline{6.13} & \underline{5.21} & \textbf{2.22}/\underline{1.11}/\underline{1.11} & 7.50 & \textbf{2.22} & \underline{2.22} & \underline{6.07} \\
+ HiLL & 26.12 & 5.63 & 4.96 & \underline{1.11}/\textbf{2.22}/\underline{1.11} & \underline{8.33} & 0.00 & \underline{2.22} & 5.74 \\
+ HATCH (ours) & \textbf{27.38} & \textbf{6.62} & \textbf{5.65} & \textbf{2.22}/\textbf{2.22}/\textbf{3.33} & \textbf{10.83} & \textbf{2.22} & \textbf{3.33} & \textbf{7.09} \\
\midrule
\textbf{Qwen3-1.7B} & 79.76 & 52.57 & 48.66 & 16.67/16.67/10.00 & 42.50 & 6.67 & 16.67 & 32.24 \\
+ GRPO & 82.57 & \underline{54.41} & 49.80 & 17.78/21.11/16.67 & 50.00 & 8.89 & \underline{26.67} & 36.43 \\
+ QuESTA & 83.17 & 54.29 & \textbf{51.49} & \underline{21.11}/18.89/\underline{17.78} & \underline{54.17} & 10.00 & 22.22 & 37.01 \\
+ HiLL & \underline{83.23} & 53.68 & \underline{50.79} & 20.00/\underline{22.22}/\underline{17.78} & 53.33 & \underline{11.11} & 22.22 & \underline{37.15} \\
+ HATCH (ours) & \textbf{83.97} & \textbf{55.02} & \textbf{51.49} & \textbf{23.33}/\textbf{25.56}/\textbf{18.89} & \textbf{58.33} & \textbf{14.44} & \textbf{28.89} & \textbf{39.99} \\
\midrule
\textbf{Qwen3-8B} & 83.57 & 57.72 & 48.36 & 26.67/20.00/20.00 & 62.50 & 6.67 & 16.67 & 38.02 \\
+ GRPO & 86.37 & 59.07 & 55.26 & 37.78/31.11/31.11 & \underline{74.17} & \underline{16.67} & 36.67 & 47.58 \\
+ QuESTA & 87.84 & \underline{61.76} & \underline{58.38} & \underline{47.78}/32.22/\underline{32.22} & 72.50 & 10.00 & 38.89 & 49.07 \\
+ HiLL & \underline{88.58} & 58.33 & 58.09 & 41.11/\underline{34.44}/31.11 & 73.33 & \underline{16.67} & \underline{43.33} & \underline{49.44} \\
+ HATCH (ours) & \textbf{90.71} & \textbf{63.11} & \textbf{60.27} & \textbf{52.22}/\textbf{38.89}/\textbf{35.56} & \textbf{80.83} & \textbf{17.78} & \textbf{44.44} & \textbf{53.76} \\
\bottomrule
\end{tabular}}
\end{center}
\end{table}

\section{Experiments}

\subsection{Experimental Setup}

\paragraph{Training settings.}
We train Llama-3.2-1B-Instruct~\citep{meta2024llama32}, Qwen3-1.7B~\citep{qwen3technicalreport}, and Qwen3-8B~\citep{qwen3technicalreport}. Within each
model scale, all methods are initialized from the same pretrained checkpoint.
We sample 10,000 problems from NuminaMath~\citep{numina_math_datasets}, spliting into 9,800 training and 200 validation set.
Training uses \textsc{verl} framework~\citep{sheng2024hybridflow}, with Megatron~\citep{megatron-lm}
backend for distributed policy optimization and vLLM~\citep{kwon2023efficient} for asynchronous rollout
generation. We use GRPO with \(G=8\) rollouts per query. For each
solve-none query, our method generates \(M=4\) hints from the current policy
and samples \(G=8\) hinted re-solves under each hint. The maximum prompt and generation lengths are 4,096 and 8,192
tokens, respectively.
Solving rollouts use temperature \(0.85\) and
top-\(p=1.0\), while hint generation uses temperature \(0.3\) and
top-\(p=0.95\). 
We use a batch size of 128 and optimize the policy with Adam~\citep{kingma2017adammethodstochasticoptimization} using a
learning rate of \(1\times10^{-6}\), a 5-step warmup and a KL penalty of \(0.01\). Experiments are conducted on 4 nodes with 8
NVIDIA H20 GPUs each.

\paragraph{Baselines.}
Rows labeled with model names in Table~\ref{tab:main-results}
report initial accuracy.
\emph{GRPO}~\citep{shao2024deepseekmath}
directly optimizes the original queries using group-relative advantages.
\emph{QuESTA}~\citep{li2025questa} is reimplemented following its two-stage curriculum, which
gradually reduces the amount of solution guidance. For a
matched online comparison, \emph{HiLL}~\citep{xia2026learning} learns utility-scored hints and trains on
the best hint.
\emph{Ours} uses the hint annealed self-teaching described in
Section~\ref{sec:method}.

\paragraph{Evaluation settings.}
For each run, we select the checkpoint with the highest no-hint
validation accuracy and report its no-hint accuracy on nine benchmarks. We report pass@1, averaged over three independent runs
with different random seeds in Table~\ref{tab:main-results}. The sample standard deviation over the three runs is at most
\(0.8\)~pp. We evaluate on Math500~\citep{lightman2023lets}, Minerva Math~\citep{lewkowycz2022solvingquantitativereasoningproblems},
OlympiadBench~\citep{he2024olympiadbenchchallengingbenchmarkpromoting}, AIME 2024--2026~\citep{aime24,aime25,aime26}, AMC 2023~\citep{AoPS:AMCProblemsSolutions}, HMMT 2025~\citep{dekoninck2026matharena}, and BRUMO 2025~\citep{dekoninck2026matharena}. Among these benchmarks, HMMT 2025 and BRUMO 2025 provide additional tests of generalization across competition-specific problem sets.

\subsection{Main Results}

Table~\ref{tab:main-results} reports original-query pass@1 across three model
scales. Our method achieves the highest macro average at every scale, improving
over GRPO by 1.70~pp on Llama-3.2-1B-Instruct, 3.56~pp on Qwen3-1.7B,
and 6.18~pp on Qwen3-8B. It also outperforms both hint-based baselines at
all three scales. The improvement is broad rather than concentrated on a single
benchmark: on Qwen3-8B, our method obtains the best score on every benchmark
in the evaluation suite, including HMMT 2025 and BRUMO 2025, which assess generalization
across different competition-specific problem distributions. These results show that the recovered trajectories improve the policy's
ability to solve the original queries, rather than merely improving performance
under hinted inputs.

\begin{figure*}[t]
\centering
\begin{minipage}[b]{0.325\textwidth}
\centering
\includegraphics[width=\linewidth]{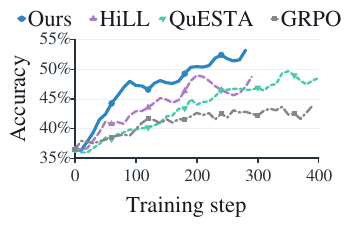}
\end{minipage}\hfill
\begin{minipage}[b]{0.325\textwidth}
\centering
\includegraphics[width=\linewidth]{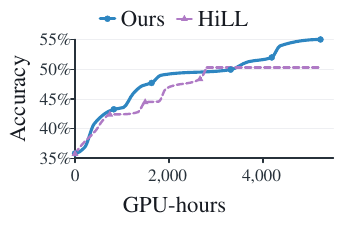}
\end{minipage}
\hfill
\begin{minipage}[b]{0.325\textwidth}
\centering
\includegraphics[width=\linewidth]{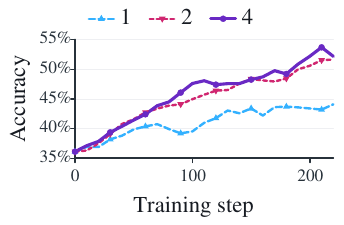}
\end{minipage}
\caption{
Left: average benchmark accuracy over training updates. Middle: best-so-far no-hint benchmark accuracy against cumulative
GPU-hours for the two online methods. Right: accuracy of training by our method with single-, two-, and four retained hinted-solve groups.
}
\label{fig:learning-dynamics}
\end{figure*}

Figure~\ref{fig:learning-dynamics} traces how the endpoint gains emerge during
training and how they depend on the number of retained hinted-solve groups on Qwen3-8B.
In the left panel, our method surpasses \(50\%\) accuracy by around
step 200 and reaches approximately \(53\%\) near step 280, whereas HiLL,
QuESTA, and GRPO remain below \(49\%\).
The middle panel provides the corresponding compute view: up to approximately
\(49\%\) best-so-far accuracy, our method reaches each level with
lower cumulative GPU-hours than HiLL.
The right panel reports the trajectory-count ablation: by step 220, the
two- and four-group settings reach approximately \(51\%\) accuracy, whereas
the single-group setting remains near \(44\%\).
Together, these dynamics show that our method achieves stronger no-hint
learning while making effective use of the additional trajectories recovered
under hints.

\subsection{Further Analysis}

\begin{wraptable}[16]{r}{0.50\textwidth}
\caption{Ablation results on Qwen3-8B.}
\label{tab:ablation}
\begin{center}
\small
\setlength{\tabcolsep}{3pt}
\begin{tabular*}{\linewidth}{@{\extracolsep{\fill}}l@{}l@{\qquad}r@{}l@{}}
\toprule
& Variant & Acc. (\%) & \\
\midrule
& HATCH (ours) & \textbf{53.76} & \\
\midrule
& Offline Hints & 48.07 & \\
& External Hints & 48.80 & \\
& Online Hints & 50.16 & \\
\midrule
& Fixed Weighting & 44.96 & \\
& Linear Decay & 48.25 & \\
& Staged Exp. Decay & 51.25 & \\
\midrule
& No Projection & 51.93 & \\
& Reverse Projection & 49.24 & \\
\bottomrule
\end{tabular*}
\end{center}
\end{wraptable}

\textbf{Improvement from single policy joint training.}
To assess the value of jointly learning hint generation and query solving, we compare our method with three variants in Table~\ref{tab:ablation}. The \emph{Offline Hints} and \emph{External Hints} variants use
fixed hints generated offline by Qwen3-8B and Qwen3-30B-A3B,
respectively. These variants achieve \(48.07\%\) and \(48.80\%\), suggesting that a larger
external hinter alone does not reproduce the gains from joint learning. To isolate the benefit of learning hint construction, \emph{Online Hints}
retains current-policy hints but removes
\(\mathcal{L}_{\mathrm{hint\mbox{-}gen}}\).
It achieves \(50.16\%\), 3.60~pp below ours. Together, these results support learning from hint construction,
beyond using hints solely as auxiliary inputs.

\begin{figure*}[!t]
\centering
\begin{minipage}[b]{0.325\textwidth}
\centering
\includegraphics[width=\linewidth]{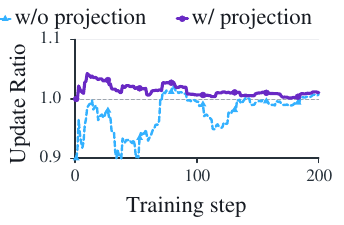}
\end{minipage}\hfill
\begin{minipage}[b]{0.325\textwidth}
\centering
\includegraphics[width=\linewidth]{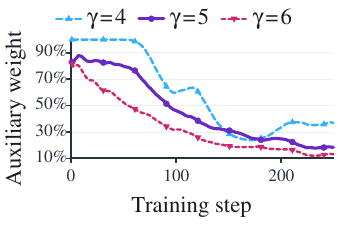}
\end{minipage}\hfill
\begin{minipage}[b]{0.325\textwidth}
\centering
\includegraphics[width=\linewidth]{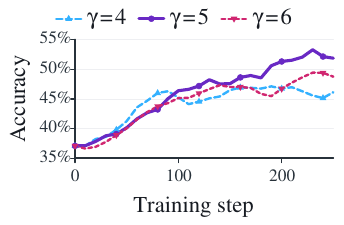}
\end{minipage}
\caption{Ablation dynamics on Qwen3-8B. 
Left: relative strength in the solution direction in the final shared update gradient.
Middle: auxiliary-weight schedules induced by the three \(\gamma\) settings.
Right: the corresponding nine-benchmark accuracy curves of each selected \(\gamma\).
}
\label{fig:ablation}
\end{figure*}

\textbf{Hinted Reward Shift Correction by Online Weighting.}
We compare data-driven hint annealing with fixed weighting and
two predefined decay schedules in Table~\ref{tab:ablation}.
Fixed weighting reduces accuracy to \(44.96\%\).
Linear decay and staged exponential decay achieve \(48.25\%\) and
\(51.25\%\).
These results support the effectiveness of data-driven hint annealing
relative to the fixed weighting and predefined decay schedules.
The middle panel of Figure~\ref{fig:ablation} shows how online
weighting reduces hinted-solve contributions as original reward
contrast emerges.
The right panel shows that \(\gamma=5\) best balances this trade-off:
annealing too quickly weakens useful early feedback, whereas annealing too
slowly sustains hinted reward shift.
Together, these results support hint annealing as a way to exploit assistance early without allowing it to dominate
once original rollouts become informative.

\textbf{Beneficial hint learning with gradient projection.}
To examine whether gradient projection helps hint-generation learning
improve query solving, we analyze its effect on solution-side updates
and final accuracy. The left panel of
Figure~\ref{fig:ablation} shows that projection preserves the shared update in
the solution direction, whereas the unprojected update is repeatedly weakened
by the conflicting component. Correspondingly, Table~\ref{tab:ablation} shows that removing projection reduces
accuracy to \(51.93\%\). Reverse projection, which retains only the opposing
component, performs substantially worse at \(49.24\%\). These results show that
the gain comes from preserving directionally compatible hint-generation
learning, rather than allowing it to counteract original-query solving.

\begin{figure*}[!t]
\centering
\begin{minipage}[b]{0.325\textwidth}
\centering
\includegraphics[width=\linewidth]{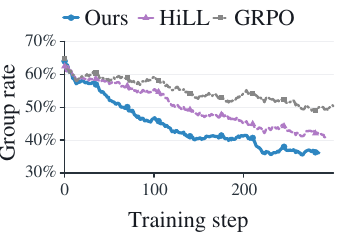}
\end{minipage}\hfill
\begin{minipage}[b]{0.325\textwidth}
\centering
\includegraphics[width=\linewidth]{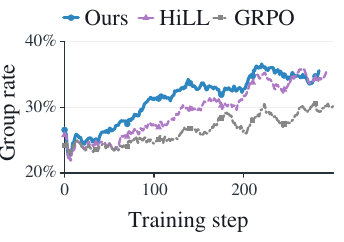}
\end{minipage}\hfill
\begin{minipage}[b]{0.325\textwidth}
\centering
\includegraphics[width=\linewidth]{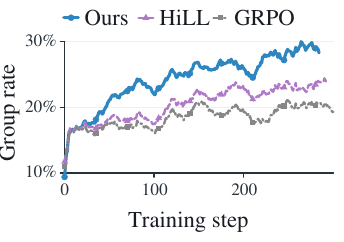}
\end{minipage}
\caption{
No-hint group-state dynamics across Ours, HiLL, and GRPO on Qwen3-8B. Ours progressively
converts solve-none groups into solve-partial and solve-all groups, indicating
stronger unassisted solving. From left to right panels are: solve-none, solve-partial and solve-all groups.
}
\label{fig:training-dynamics}
\end{figure*}

\textbf{Improved query solving.}
Recovered hinted trajectories should improve how the policy solves a query
itself, rather than only its behavior after receiving a hint. To examine this,
Figure~\ref{fig:training-dynamics} tracks the composition of rollout groups throughout training. Compared with other methods,
our method reduces the fraction of solve-none groups more rapidly while
increasing solve-partial and solve-all groups. This
shows that we better improve solving behavior beyond the
hinted condition.

\subsection{Limitations and Future Work}
First, our evaluation focuses on mathematical reasoning with verifiable
rewards and reference. In interactive tasks such as tool use, new observations can change the
guidance needed at each step. Extending HATCH to
such tasks therefore requires verification and reference
suited to the context.
Second, our experiments use only textual inputs and hints. Future work
could explore multimodal self-teaching by incorporating visual evidence
into hint construction and subsequent reasoning.

\section{Conclusion}
We discover \emph{hinted reward shift}, where recovered signal can favor hinted solving without corresponding gains without hints.
To address this, we propose \textsc{HATCH} (\emph{Hint-Annealed
Self-Teaching}), which jointly learns hint generation and query solving within one policy, using online weighting and gradient projection to regulate the strength and direction of auxiliary learning.
Experiments across three model scales show stronger unassisted
mathematical reasoning and suggest that models can improve their reasoning by learning to provide effective guidance for themselves.

\section*{Full Author List}
\label{sec:author-list}
Zile Wang, Zijian Li, Haodong Wang, Jian Liu, Qianli Liu, Lucas Muli,
Blaze Chen, Song Guo

\bibliography{ref}
\bibliographystyle{uxbench}

\end{document}